\documentclass[11pt]{article}

\usepackage[final]{acl}

\usepackage{times}
\usepackage{latexsym}
\usepackage[T1]{fontenc}
\usepackage[utf8]{inputenc}
\usepackage{microtype}
\usepackage{inconsolata}

\usepackage{pifont}
\newcommand{\cmark}{\ding{51}}
\usepackage{multirow}
\usepackage{graphicx}
\usepackage{subcaption}
\usepackage{wrapfig}
\usepackage{makecell}
\usepackage{tcolorbox}
\tcbuselibrary{breakable}
\usepackage{xcolor}
\usepackage{enumitem}
\usepackage{booktabs}
\usepackage{amssymb}
\usepackage[table]{xcolor}
\definecolor{promptbg}{HTML}{FBFBFB}
\definecolor{promptborder}{HTML}{D1D5DB}
\setitemize[1]{itemsep=0pt,partopsep=0pt,parsep=0pt,topsep=0pt,leftmargin=12pt}
\definecolor{header_gray}{RGB}{230, 230, 230}
\definecolor{emerald}{RGB}{3, 115, 66}
\definecolor{zhu}{RGB}{209, 41, 32}
\definecolor{new_blue}{RGB}{62, 70, 230}

\title{Specializing Large Models for Oracle Bone Script Interpretation via Component-Grounded Multimodal Knowledge Augmentation}

\author{
  Jianing Zhang$^{1, *}$,
  Runan Li$^{1, *}$,
  Honglin Pang$^{2}$,
  Ding Xia$^{3}$, \\
  \textbf{Zhou Zhu}$^{4}$,
  \textbf{Qian Zhang}$^{4}$,
  \textbf{Chuntao Li}$^{4, 5, \dagger}$,
  \textbf{Xi Yang}$^{2, 5, \dagger}$ \\
  $^1$College of Software, Jilin University \quad
  $^2$School of Artificial Intelligence, Jilin University \\
  $^3$Graduate School of Information Science and Technology, The University of Tokyo \\
  $^4$School of Archaeology, Jilin University \\
  $^5$Engineering Research Center of Knowledge-Driven Human-Machine Intelligence, MoE, China \\
  $^*$Equal contribution \quad $^\dagger$Corresponding author \\
  \texttt{\{zhangjn5523,panghl25\}@mails.jlu.edu.cn} \quad \texttt{\{dingxia1995,earthyangxi\}@gmail.com} \\
}

\begin{document}
\maketitle
\begin{abstract}

Deciphering ancient Chinese Oracle Bone Script (OBS) is a challenging task that offers insights into the beliefs, systems, and culture of the ancient era. Existing approaches treat decipherment as a closed-set image recognition problem, which fails to bridge the ``interpretation gap'': while individual characters are often unique and rare, they are composed of a limited set of recurring, pictographic components that carry transferable semantic meanings. To leverage this structural logic, we propose an agent-driven Vision-Language Model (VLM) framework that integrates a VLM for precise visual grounding with an LLM-based agent to automate a reasoning chain of component identification, graph-based knowledge retrieval, and relationship inference for linguistically accurate interpretation.
To support this, we also introduce OB-Radix, an expert-annotated dataset providing structural and semantic data absent from prior corpora, comprising 1,022 character images (934 unique characters) and 1,853 fine-grained component images across 478 distinct components with verified explanations. By evaluating our system across three benchmarks of different tasks, we demonstrate that our framework yields more detailed and precise decipherments compared to baseline methods.
\end{abstract}

\section{Introduction}

\begin{figure}[t]
  \centering
  \includegraphics[width=0.75\linewidth]{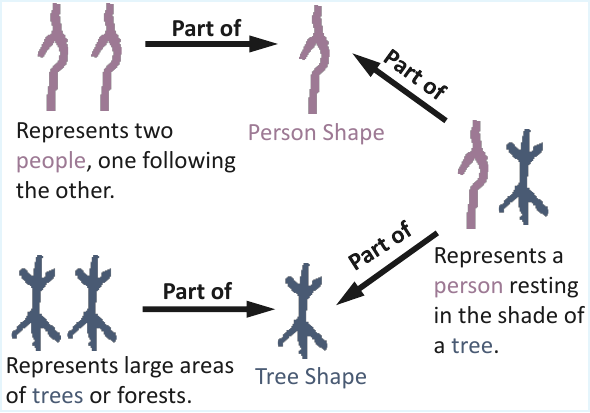}
  \caption{Oracle Bone Script (OBS), a pictographic writing system of semantic components.}
  \label{fig:xiu}
\end{figure}

Oracle Bone Script (OBS), the earliest known mature writing system in China, holds significant historical and cultural value. Of the more than 4,500 identified OBS characters, only approximately one-third have been deciphered, leaving a vast corpus of glyphs in mystery \citep{li2024oracleboneinscriptionsmultimodal}. Each undeciphered character represents a lost fragment of ancient institutions, technologies, and beliefs. However, the fragmented and stylised nature of OBS inscriptions, coupled with the requirement for profound paleographic and contextual expertise, renders manual decipherment exceptionally difficult.

In recent years, artificial intelligence has increasingly been leveraged for OBS interpretation \citep{fu2022improvement,wang2024puzzle,guan-etal-2024-deciphering,jiang2023oraclepoints}. Most existing methods treat decipherment as a closed-set image recognition task, largely neglecting the structural, semantic, and contextual nuances intrinsic to the script. This narrow focus results in information waste and introduces interpretive biases, as models lack the domain-specific knowledge required to generalise beyond known characters. While Vision-Language Models (VLMs) have demonstrated robust general image–text understanding \citep{liu2023visual,caffagni-etal-2024-revolution}, their capacity for fine-grained perception and expert reasoning remains a bottleneck. In low-resource, specialised domains like OBS, standard VLMs often suffer from ``visual hallucinations'' or a lack of linguistic depth, leading to bad performance \citep{chen2025obibench,ye2024exploring}.

\begin{figure*}[t]
  \centering
  \includegraphics[width=0.85\textwidth]{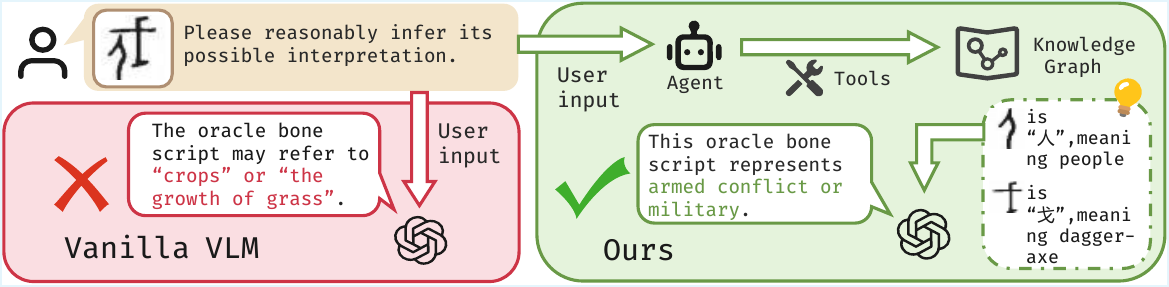}
  \caption{Comparison of our proposed framework and baselines. We design an agentic RAG framework to integrate component-level knowledge for structured semantic augmentation of OBS.}
  \label{teaser}
\end{figure*}

Structurally, the OBS is an organised system of pictographic components, where each character is made up of discrete radicals that carry a distinct semantic weight and are frequently reused throughout the lexicon (Figure~\ref{fig:xiu}).  This component-based architecture provides a critical logic bridge for decipherment: by identifying known pictographic components within an unknown glyph, we can systematically infer the meaning of characters that are new to the model. To leverage this, we propose an Agentic Retrieval-Augmented Generation (Agentic RAG) framework that empowers VLMs with component-based semantic augmentation (Figure~\ref{teaser}). To support this, we introduce OB-Radix, a new expert-annotated dataset comprising 1,022 Oracle character images (934 unique characters) and 1,853 fine-grained component images (478 distinct components), each paired with expert-verified semantic explanations. Finally, to evaluate whether our approach achieves expert-level capability, we design three progressively advanced benchmarks: (1) component-level retrieval, (2) component relationship inference, and (3) OBS interpretation generation. Experimental results demonstrate that our framework outperforms baseline methods, providing a more interpretable and linguistically accurate pathway for OBS decipherment. 

In summary, our contributions are:

\begin{itemize}
    \item We reformulate oracle bone script (OBS) interpretation as a \emph{component-grounded, structure-aware reasoning task}, rather than a purely visual recognition problem, and instantiate this formulation with a multimodal framework that integrates component-level visual cues and graph-based retrieval.
    
    \item We construct OB-Radix, a component-level oracle bone script dataset, and build a knowledge graph that captures relationships among components, characters, and their semantic explanations, providing essential structured knowledge.
    
    \item We design comprehensive evaluations to assess both the accuracy and interpretability of our approach. Results show that our framework produces interpretations closely aligned with expert annotations and that the multi-agent extension offers enhanced semantic grounding.

\end{itemize}

\section{Related Work}
\textbf{Deciphering of Oracle Bone Script.} 
Existing research relies on a single image morphology model to explore AI reading paths. \cite{guan-etal-2024-deciphering} employs a diffusion approach to map oracle bone inscription images to modern Chinese characters, while \cite{qiao2024making} leverages image generation to provide visual interpretive guidance. However, the former lacks integration of textual semantics, and the latter results in incomplete understanding due to the absence of textual guidance. Other studies applied diverse AI techniques \citep{fu2022improvement,jiang2023oraclepoints,wang2024puzzle,gan2023characters} from different perspectives to aid in the decipherment of Oracle Bone Script. 

\textbf{Graph Retrieval-Augmented Generation for VLMs.} Although large-scale VLMs demonstrate strong zero-shot generalization, they still exhibit noticeable performance drops when the underlying training corpora lack or misrepresent the necessary domain knowledge \citep{zhang2024vision,minaee2024large}.
To enhance the specialization of visual language approaches in particular domains, Retrieval-Augmented Generation (RAG) approaches are employed \citep{lin2024revolutionizing,zhang2025plantgpt}. Unlike traditional fine-tuning, RAG dynamically retrieves relevant knowledge from external databases during inference, enabling VLMs to access domain-specific information on-demand without updating their pre-trained parameters.
Additionally, to mitigate the potential noise present in general knowledge bases that may affect results, the concise representation provided by knowledge graphs are integrated, forming what is known as Graph RAG \citep{peng2024graph}.

\textbf{Oracle Bone Script Datasets.}
Most existing oracle bone script (OBS) datasets focus on \emph{character-level} recognition, providing complete character images for end-to-end modeling, such as HUST-OBC \citep{wang2024open}, EVOBC \citep{guan2024opendatasetevolutionoracle}, OBC306 \citep{huang2019obc306}, Oracle-50k \citep{han2020self}, and HWOBC \citep{li2020hwobc}. While these datasets cover multiple historical scripts, they lack \emph{component-level} annotations and therefore provide limited support for structural decomposition and interpretable semantic analysis.
OracleFusion \citep{li2025oraclefusion} introduces radical-level structures, bounding boxes, and semantic concepts for oracle characters, but its annotations remain region-based and lack expert-curated component entities, consistent semantic interpretations, and explicit inter-component relations, limiting its support for component-grounded reasoning and knowledge graph construction.

\section{OB-Radix Dataset}

\begin{figure}[t]
  \centering
  \includegraphics[width=\columnwidth]{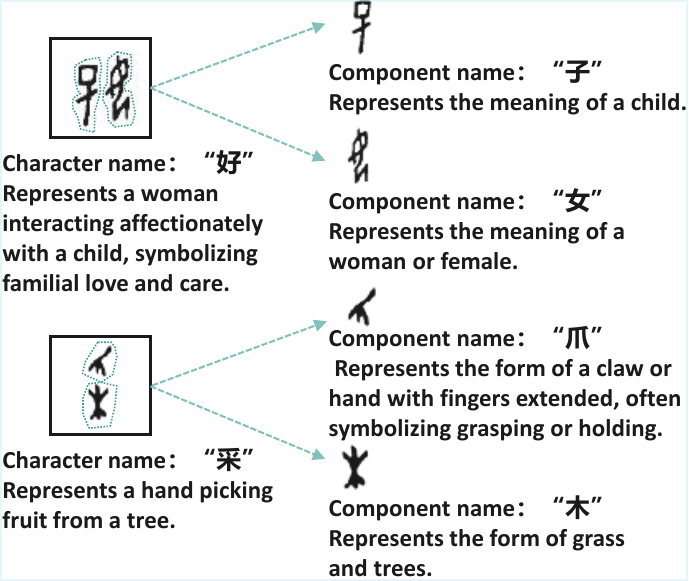}
  \caption{Our annotation of an oracle character at the component level.}
  \label{annot}
\end{figure}

We introduce OB-Radix, a dataset of hierarchical structural relations and grounded visual data meticulously curated by experts in paleography. Unlike prior character-level datasets \citep{wang2024open,guan2024opendatasetevolutionoracle,huang2019obc306,han2020self,li2020hwobc} or those relying on ungrounded visual fragments \citep{hu2024component} and text-only decompositions \citep{jiang2024oraclesage}, OB-Radix shifts the focus from the whole character to its constituent parts. This design enables models to learn and utilize the underlying compositional logic of the script, facilitating the interpretation of previously unseen or undeciphered glyphs through the identification of known pictographic elements. OB-Radix comprises 1,022 oracle character images covering 934 unique characters, along with 1,853 fine-grained component images spanning 478 distinct components.

Three archaeology doctoral students were tasked with identifying components based on their paleographic function and meaning, rather than relying solely on stroke continuity or visual salience. Specifically, annotators followed three core principles: (i) isolating components with distinct semantic roles, regardless of their visual scale; (ii) prioritizing semantic integrity over geometric completeness in cases of ambiguous boundaries; and (iii) maintaining uniform component labels across the corpus through a controlled vocabulary. These principles ensure that each component serves as a reliable semantic anchor, directly mapping visual regions to specific entries in our paleographic knowledge base for downstream reasoning.

Figure~\ref{annot} showcases representative examples of this expert-level annotation, illustrating the decomposition of OBS characters into semantically meaningful components rather than arbitrary visual regions. To achieve high-precision segmentation, we utilized LabelMe \citep{russell2008labelme} to perform semantic masking. Annotators were required to delineate the component region, after which the software automatically masked the largest contiguous region as the component body. To further refine the quality, experts manually adjusted key boundary points on each mask, ensuring the segmentation precisely captures the pictographic structure of the OBS. Given the specialized expertise required for such tasks, the curation process involved 70 total man-hours, representing a significant investment in high-fidelity data for the OBS domain. Further implementation details regarding the annotation tool, the expert workflow, and quality control measures are provided in Appendix \ref{details_dataset}.

\begin{figure*}[t]
  \centering
  \includegraphics[width=0.99\textwidth]{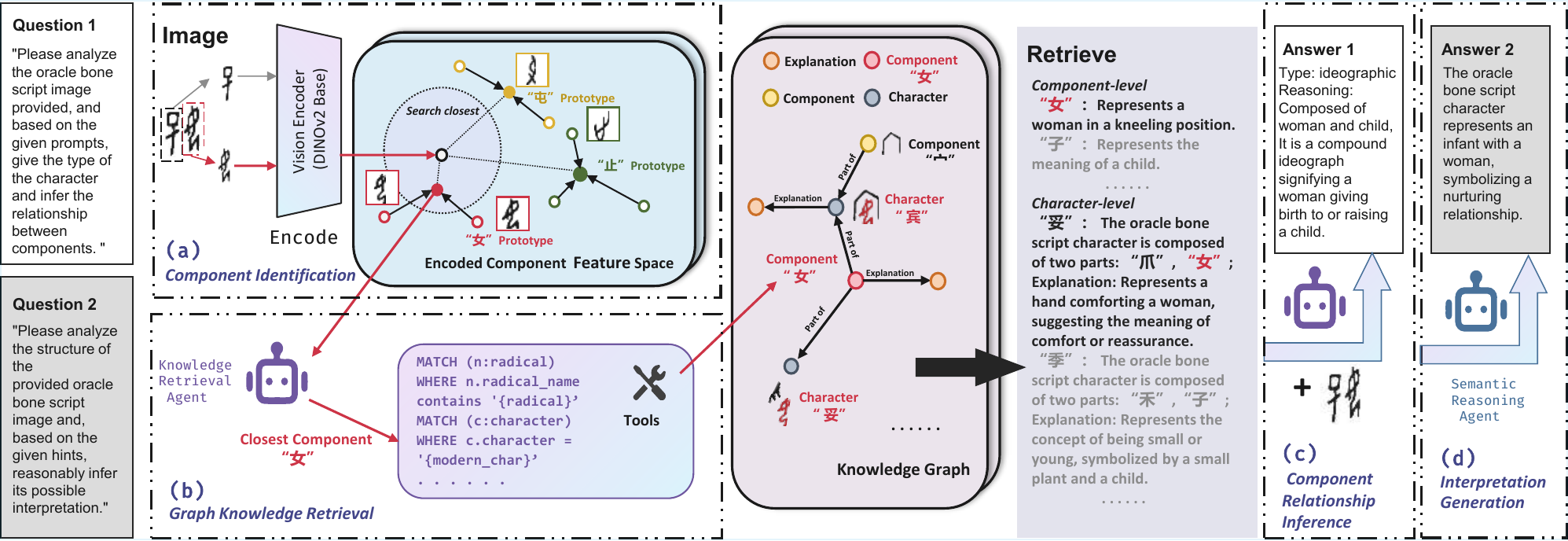}

  \begin{subfigure}{0.24\textwidth}
    \phantomcaption
    \label{fig:stage_a}
  \end{subfigure}
  \begin{subfigure}{0.24\textwidth}
    \phantomcaption
    \label{fig:stage_b}
  \end{subfigure}
  \begin{subfigure}{0.24\textwidth}
    \phantomcaption
    \label{fig:stage_c}
  \end{subfigure}
  \begin{subfigure}{0.24\textwidth}
    \phantomcaption
    \label{fig:stage_d}
  \end{subfigure}

  \caption{Detailed pipeline of our approach: (a) Component Identification Module identifies radical components from input OBS images; (b) Agent-Driven Graph Knowledge Retrieval retrieves relevant information from our constructed knowledge graph; (c) Component Relationship Inference uses VLMs to determine the structural relationships among components; (d) Interpretation Generation produces comprehensive semantic interpretations of oracle characters.}
  \label{fig:p3}
\end{figure*}


\section{Method}
As shown in Figure~\ref{fig:p3}, our approach integrates visual analysis of OBS with structured knowledge reasoning through an agent-driven retrieval-augmented generation pipeline, comprising four parts: (1) a component identification module through character radicals retrieval as shown in Figure~\ref{fig:stage_a}, (2) an agent-driven knowledge graph retrieval module to dynamically query relevant entries as shown in Figure~\ref{fig:stage_b}, (3) a component relationship analysis and judgment module as shown in Figure~\ref{fig:stage_c} and (4) an interpretation generation module that integrates full character-level explanations as shown in Figure~\ref{fig:stage_d}. And the interpretation module supports two inference strategies: a VLM-based mode that directly fuses visual features with retrieved knowledge, and a multi-agent mode that separates retrieval and reasoning into specialized agents, enhancing robustness and interpretability.

\subsection{Component Identification}\label{sec:component_identification}
To identify radical components from input OBS images, we first utilize a Vision Transformer (ViT) architecture based on DINOv2 \citep{dosovitskiy2021an,oquab2024dinov} to construct a component feature space, as it produces highly transferable features. Then, we adopt a prototype-based classifier following Prototypical Networks \citep{snell2017prototypical}, as its class-level aggregation is well-suited to our low-data regime, improving robustness and reducing overfitting.

Specifically, Eq.~(1) defines how the input radical image $\mathbf{x}$ is encoded by the DINOv2 encoder $f(\cdot)$ into a 768-dimensional vector $\mathbf{z}$, and we then compute its prototype $\mathbf{p}_c$ as the mean embedding of its support set $\mathcal{S}_c$ for each class $c$. Thus, given a query image $\mathbf{x}_q$, its embedding $\mathbf{z}_q = f(\mathbf{x}_q)$ is compared to all class prototypes using Euclidean distance $d(\cdot,\cdot)$, and then classified into the class with the nearest prototype.
\begin{equation}
\mathbf{z} = f(\mathbf{x}), \mathbf{z} \in \mathbb{R}^{768}
\end{equation}
\begin{equation}
\hat{y} = \arg\min_{c} d\bigl(\mathbf{z}_q, \mathbf{p}_c\bigr)
\end{equation}

Compared with directly using conventional classifiers or detectors, this design enables our model to make efficient use of limited labeled samples and enhances generalization in the low-resource setting of OBS component identification.
An illustrative visualization of the component feature space construction is provided in Appendix~\ref{app:feature_space}.

\subsection{Agent-Orchestrated Graph Knowledge Retrieval}

We construct a Knowledge Graph (KG) from OB-Radix and character–component relations. For each test character, the PrototypeClassifier first predicts its most likely components; these predicted components are then used as \emph{primary semantic cues} to query the KG. Rather than learning an unconstrained policy, we adopt a \textbf{cascading but largely fixed} retrieval pipeline, orchestrated by a tool-using LLM agent \citep{yao2023react,schick2023toolformer}. The agent can call two external tools—\emph{component explanation} and \emph{characters-by-component}—and performs additional reasoning internally. Concretely:

\begin{itemize}
\item \textit{Component-centric retrieval.} Given the predicted components, the agent first queries their explanations and searches for characters that contain these components, which typically provide the most direct semantic evidence.
\item \textit{Constrained synthesis.} When component-based retrieval yields weak or insufficient evidence, the agent internally performs variant lookup and modern–oracle mapping—without invoking external tools—to supplement the retrieved information. It then summarizes and reorders all evidence, both tool-obtained and internally inferred, into a concise, character-centric evidence bundle as input to the interpretation module.
\end{itemize}

To improve efficiency, we integrate a simple semantic-similarity cache following \citet{jin2024ragcache}, so that repeated or near-duplicate KG queries are served from cache. Overall, the agent acts as a lightweight orchestration layer over a deterministic retrieval cascade, ensuring that knowledge access is predictable and efficient while still providing rich, component-grounded context for downstream interpretation.

\subsection{Component Relationship Inference}
To move beyond black-box recognition, we design a module that leverages VLMs to infer the structural relationships among components. After the components are identified and the knowledge graph retrieval refines them, the system uses a VLM to jointly consider both visual embeddings and retrieved semantic information. The task requires the model to predict the inscription type of each oracle character, which can be categorized as ideographic, pictographic, or phono-semantic, and to generate a reasoning trace that explains how the components interact to form meaning. This process is illustrated in Figure~\ref{fig:p3}, while the resulting output are presented in Figure~\ref{fig:stage_c}.

By conditioning the VLM on both structural and semantic cues, the module produces explanations that are not only accurate but also interpretable to human users. The component-level information is integrated into reasoning about character structure and provides the intermediate reasoning layer that connects recognition and interpretation generation.

\subsection{Interpretation Generation}

To generate full semantic interpretations of oracle characters, we design an inference pipeline that integrates visual recognition with knowledge-graph-based reasoning. Our framework supports two complementary modes of inference.

The first mode, \emph{VLM Inference}, employs a VLM that jointly conditions on the visual embeddings of the inscription, component predictions from the PrototypeClassifier, and semantic prompts retrieved from the knowledge graph. By grounding ambiguous visual forms in curated historical evidence, the VLM produces interpretations that are semantically coherent and visually faithful.

Building upon this design, we further introduce a second mode, \emph{Multi-Agent Inference}, inspired by recent advances in cooperative agent systems \citep{wu2023autogen, chang2024main, jin-etal-2025-disentangling-memory, nguyen2025ma, singh2025agentic, wu-etal-2025-agentic}. We use multi-agents to decouple retrieval and reasoning functions. A \emph{Knowledge Retrieval Agent} plans and executes graph queries to gather relevant evidence, while a \emph{Semantic Reasoning Agent} synthesizes this evidence with visual cues into structured, human-interpretable explanations. This separation improves robustness, reduces error propagation, and leverages the natural ability of large models to think after retrieval.

\section{Experiments}

To systematically evaluate whether our approach achieves expert-level capability in OBS interpretation, we design a series of experiments under expert guidance, structured around three progressively advanced tasks: (1) component-level retrieval as the foundation, (2) component relationship inference as the intermediate stage, and (3) OBS interpretation generation as the ultimate goal.

\subsection{Metrics and baselines}
We report ACC@k ($k \in \{1, 3, 5\}$) for component retrieval, and the accuracy of the oracle-character type classification for the component relationship inference experiment. We employ BERTScore-F1, MoverScore, ROUGE-1, and an LLM-as-a-Judge paradigm for OBS interpretation \citep{Zhang*2020BERTScore:,zhao-etal-2019-moverscore,lin-2004-rouge,zheng2023judging}. To ensure evaluation impartiality \citep{li2025preference}, we instantiate the judge using Gemini~3~Flash \citep{team2023gemini}. Details of the LLM-as-a-Judge setup, including the evaluation rubric, prompting strategy, and the 0--1 scoring scale, are provided in Appendix~\ref{app:judge}.

In the experimental tables, we use shorthand notations for VLMs. Specifically, \textit{GPT} refers to GPT-5 \citep{2025gpt5}; \textit{Claude} refers to Claude Opus 4.1 (20250805) \citep{2025claude41}; \textit{GLM} refers to GLM-4.5V \citep{vteam2025glm45vglm41vthinkingversatilemultimodal}; and \textit{Qwen} refers to Qwen3-VL-235B-A22B \citep{qwen2025qwen3vl}.

\subsection{Dataset Splitting}
\label{exp:details}
We adopted consistent dataset splitting strategies to ensure fair and realistic evaluation for all experiments. Specifically:

\begin{itemize}
    \item \textbf{Component retrieval} (Section~\ref{exp1}): Our OB-Radix dataset, containing 478 distinct components, was divided into training and testing sets with a ratio of 7:3, respectively. Model performance was measured by Top-1, Top-3, and Top-5 accuracy.  

    \item \textbf{Component relationship inference} (Section~\ref{exp2}): We constructed a seen set of 528 annotated instances, each including both inscription type labels and expert-derived reasoning traces. Models were trained and evaluated on this split without data overlap, ensuring interpretability analysis was grounded in expert references.  

    \item \textbf{Interpretation generation} (Section~\ref{exp3}): To avoid leakage, our KG was built using 70\% of the corpus, while the remaining 30\% was held out for testing. This split applies to all experiments related to Section~\ref{exp3}. It guarantees that characters used for evaluation had not appeared in training, thus presenting a realistic challenge of interpreting previously unseen instances.
\end{itemize}

\begin{table}[t]
  \centering
  \caption{OBS component retrieval results.}
  \vspace{-6pt}
  \setlength{\tabcolsep}{6pt}
  \small
  \begin{tabular}{l c}
    \toprule
    \rowcolor{header_gray}
    \textbf{Metric} & \textbf{ACC$\uparrow$} \\
    \midrule
    Top-1 & 0.7795 \\
    Top-3 & 0.8855 \\
    Top-5 & 0.9157 \\
    \bottomrule
  \end{tabular}
  \label{tab:radical_recognition_results}
\end{table}
\subsection{Component Identification}
\label{exp1}

The most essential prerequisite for understanding oracle bone characters lies in the ability to accurately recognize their constituent components, since these components serve as the fundamental units from which higher-level semantic and structural interpretations are derived. As summarized in Table~\ref{tab:radical_recognition_results}, our approach achieves competitive recognition accuracy, demonstrating its effectiveness in capturing the visual and structural properties of OBS. 

\subsection{Component Relationship Inference}\label{exp2}

\begin{figure}[t]
  \centering
  \includegraphics[width=0.9\columnwidth]{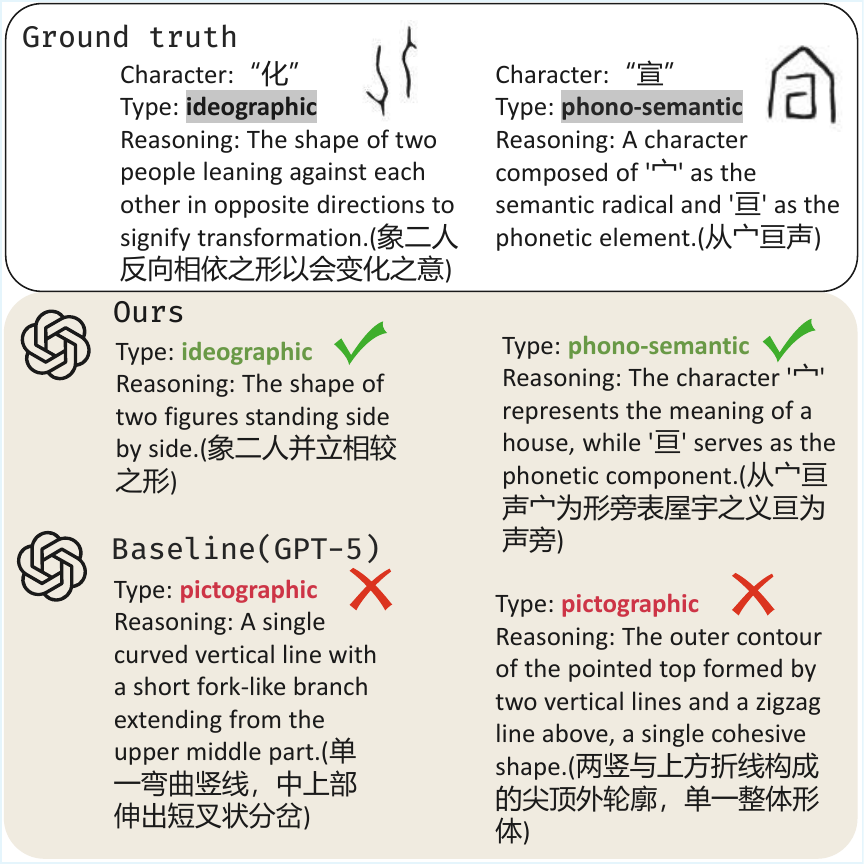}
  \caption{Reasoning examples for component relationship inference. 
\textit{Ground truth} shows expert interpretations.}
  \label{Fig:type_reasoning}
\end{figure}

We evaluate whether VLMs capture the structural relationships among components, rather than treating OBS recognition as a black-box task. The task involves: (1) predicting the inscription type of a character (ideographic, pictographic, or phono-semantic), and (2) generating a textual explanation of component interactions. Representative examples comparing baseline and our enhanced pipeline are shown in Figure~\ref{Fig:type_reasoning}.  

Table~\ref{tab:results} reports classification and reasoning results. Our component-aware pipeline outperforms baselines across all metrics, confirming that explicit component-level knowledge improves both accuracy and interpretability. Qwen3-VL achieves the highest classification accuracy (0.599), while GPT-5 achieves the best performance under both BERTScore and LLM-as-a-Judge evaluation. Claude Opus 4.1 further shows the strongest fluency and alignment in reasoning (MoverScore, ROUGE-1).


\newcommand{\imp}[1]{\textsuperscript{\textcolor{emerald}{#1}}}
\begin{table}[t]
    \centering
    \scriptsize
    \setlength{\tabcolsep}{0.9pt}
    \renewcommand{\arraystretch}{1.1} 
    \caption{OBS component relationship inference results.}
    \label{tab:results}
    \begin{tabular}{@{}lcccccc@{}}
        \toprule
        \rowcolor{header_gray}
        \textbf{Category} & \textbf{Model} & \textbf{ACC$\uparrow$} & \textbf{BERT$\uparrow$} & \textbf{Mover$\uparrow$} & \textbf{ROUGE-1$\uparrow$} & \textbf{LLM-Judge$\uparrow$} \\
        \midrule
        \multirow{4}{*}{\textit{\textbf{Baseline}}}
            & GPT    & 0.364 & 0.497 & 0.310 & 0.007 & 0.237 \\
            & Claude & 0.475 & 0.495 & 0.324 & 0.009 & 0.225 \\
            & GLM    & 0.447 & 0.519 & 0.293 & 0.010 & 0.113 \\
            & Qwen   & 0.350 & 0.503 & 0.318 & 0.012 & 0.165 \\
        \midrule
        \multirow{4}{*}{\textit{\textbf{Ours}}}
            & GPT
            & 0.563\imp{+0.199}
            & \textbf{0.670}\imp{+0.173}
            & 0.472\imp{+0.161}
            & 0.199\imp{+0.192}
            & \textbf{0.435}\imp{+0.198} \\
            & Claude
            & 0.551\imp{+0.075}
            & 0.648\imp{+0.152}
            & \textbf{0.490}\imp{+0.166}
            & \textbf{0.221}\imp{+0.212}
            & 0.412\imp{+0.187} \\
            & GLM
            & 0.468\imp{+0.021}
            & 0.606\imp{+0.088}
            & 0.440\imp{+0.148}
            & 0.139\imp{+0.129}
            & 0.262\imp{+0.149} \\
            & Qwen
            & \textbf{0.599}\imp{+0.248}
            & 0.658\imp{+0.156}
            & 0.481\imp{+0.164}
            & 0.212\imp{+0.200}
            & 0.371\imp{+0.206} \\
        \bottomrule
    \end{tabular}
\end{table}

\subsection{Interpretation Generation}
\label{exp3}
This task provides a direct test of whether the system can go beyond recognition and structural reasoning to generate semantically meaningful interpretations.

We compare two categories of approaches: (1) \textit{Baseline} models, where LLMs directly generate interpretations without access to the Knowledge Graph; and (2) \textit{Agentic RAG} (ours), where the LLM retrieves supporting evidence from the graph before generating explanations. Performance was evaluated using BERTScore, MoverScore, ROUGE-1 and LLM-as-a-Judge with higher values indicating better alignment with expert-written ground truth. A concrete illustration is provided in \ref{exp3example}. Results are shown in Table \ref{tab:model_performance}.

\begin{table}[t]
\centering
\scriptsize
\renewcommand{\arraystretch}{1.1} 
\caption{OBS interpretation generation results.}
\label{tab:model_performance}
\setlength{\tabcolsep}{1.5pt} 
\begin{tabular}{@{}lccccc@{}}
\toprule
\rowcolor{header_gray}
\textbf{Category} & \textbf{Model} & \textbf{BERT$\uparrow$} & \textbf{Mover$\uparrow$} & \textbf{ROUGE-1$\uparrow$} & \textbf{LLM-Judge$\uparrow$} \\
\midrule
\multirow{4}{*}{\textit{\textbf{Baseline}}}
& GPT     & 0.633 & 0.393 & 0.227 & 0.102 \\
& Claude  & 0.614 & 0.365 & 0.232 & 0.052 \\
& GLM     & 0.634 & 0.338 & 0.275 & 0.042 \\
& Qwen    & 0.636 & 0.362 & 0.264 & 0.076 \\
\midrule
\multirow{4}{*}{\textit{\textbf{Ours}}}
& GPT    
& \textbf{0.727}\imp{+0.094}
& 0.475\imp{+0.082}
& 0.321\imp{+0.094}
& \textbf{0.558}\imp{+0.456} \\

& Claude 
& 0.716\imp{+0.102}
& 0.474\imp{+0.109}
& 0.335\imp{+0.103}
& 0.382\imp{+0.330} \\

& GLM    
& 0.706\imp{+0.072}
& 0.453\imp{+0.115}
& 0.337\imp{+0.062}
& 0.212\imp{+0.170} \\

& Qwen   
& 0.722\imp{+0.086}
& \textbf{0.471}\imp{+0.109}
& \textbf{0.354}\imp{+0.090}
& 0.339\imp{+0.263} \\
\bottomrule
\end{tabular}
\end{table}

The results clearly indicate the benefits of retrieval-augmented generation. Across all models, the Agentic RAG pipeline consistently outperforms the baseline counterparts. For example, Qwen3-VL improves from 0.264 $\rightarrow$ 0.354 on ROUGE-1 and from 0.362 $\rightarrow$ 0.471 on MoverScore. Similarly, GPT-5 achieves the best BERTScore of 0.727 and the best LLM-as-a-Judge of 0.558 under the RAG setting, demonstrating stronger semantic alignment. These findings suggest that grounding interpretation generation in structured knowledge not only enhances factual accuracy but also produces outputs that are more coherent and interpretable.

\subsection{Ablation Study}

\newcommand{\impneg}[1]{\textsuperscript{\scriptsize{\textcolor{zhu}{#1}}}}
\begin{table}[t]
\centering
\scriptsize
\caption{OBS relationship inference results.}
\label{tab:ablation_llms}
\setlength{\tabcolsep}{1.5pt}
\begin{tabular}{@{}lccccc@{}}
\toprule
\rowcolor{header_gray}
\textbf{Model} & \textbf{Retrieval} & \textbf{BERT$\uparrow$} & \textbf{Mover$\uparrow$} & \textbf{ROUGE-1$\uparrow$} & \textbf{LLM-Judge$\uparrow$} \\
\midrule
\multirow{2}{*}{GPT}
  & \cmark 
  & \textbf{0.727}
  & \textbf{0.475}
  & 0.321
  & \textbf{0.558} \\
  &        
  & 0.717\impneg{-0.010}
  & 0.469\impneg{-0.006}
  & 0.305\impneg{-0.016}
  & 0.542\impneg{-0.016} \\
\midrule
\multirow{2}{*}{Claude}
  & \cmark 
  & 0.716
  & 0.474
  & 0.335
  & 0.382 \\
  &        
  & 0.699\impneg{-0.017}
  & 0.451\impneg{-0.023}
  & 0.286\impneg{-0.049}
  & 0.372\impneg{-0.010} \\
\midrule
\multirow{2}{*}{GLM}
  & \cmark 
  & 0.706
  & 0.453
  & \textbf{0.337}
  & 0.212 \\
  &        
  & 0.687\impneg{-0.019}
  & 0.415\impneg{-0.038}
  & 0.283\impneg{-0.054}
  & 0.180\impneg{-0.032} \\
\midrule
\multirow{2}{*}{Qwen}
  & \cmark 
  & 0.722
  & 0.471
  & 0.354
  & 0.339 \\
  &        
  & 0.711\impneg{-0.011}
  & 0.445\impneg{-0.026}
  & 0.326\impneg{-0.028}
  & 0.281\impneg{-0.058} \\
\bottomrule
\end{tabular}
\end{table}

To isolate the contribution of the Agent-Driven Graph Knowledge Retrieval, we conducted an ablation experiment in which retrieval was disabled and only component category predictions were provided. The results are summarized in Table~\ref{tab:ablation_llms}. 

Across all four models, the absence of retrieval consistently reduces performance, confirming that the Oracle Knowledge Graph supplies non-trivial semantic context beyond visual recognition and component classification. Specifically, removing retrieval consistently degrades performance across all models and metrics.
The performance drops are most evident on ROUGE-1 and LLM-as-a-Judge, indicating that Agent-Driven Graph Knowledge Retrieval provides crucial relational and contextual information beyond component category predictions.
Moreover, the larger reductions in LLM-as-a-Judge compared to embedding-based metrics suggest that retrieval primarily improves higher-level semantic correctness rather than surface-level similarity.
These results confirm that graph-based knowledge retrieval is essential for reliable OBS relationship inference.

\subsection{Multi-Agent Collaboration}
\begin{figure*}[t]
  \centering
  \includegraphics[width=0.99\textwidth]{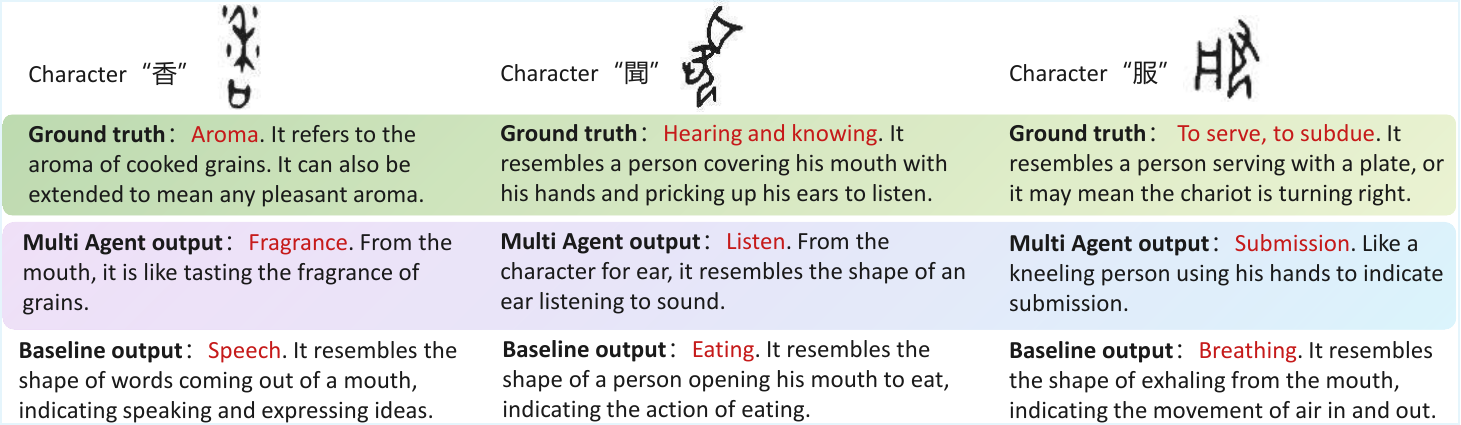}
  \caption{
  Comparison of approach outputs.
  \textit{Character} displays the original Oracle bone characters;
  \textit{Ground truth} provides the ground truth interpretations;
  \textit{Multi-agent output} shows our multi-agent approach using Graph RAG;
  \textit{Baseline output} presents results from the baseline approach.
  }
  \label{fig:instance}
\end{figure*}

\begin{table}[t]
\centering
\scriptsize
\caption{Performance comparison of multi-agent configuration.}
\label{tab:multi_agent}
\setlength{\tabcolsep}{1.5pt}
\begin{tabular}{@{}llcccc@{}}
\toprule
\rowcolor{header_gray}
\textbf{Retriever} & \textbf{Reasoner} & \textbf{BERT$\uparrow$} & \textbf{Mover$\uparrow$} & \textbf{ROUGE-1$\uparrow$} & \textbf{LLM-Judge$\uparrow$} \\
\midrule
\multirow{3}{*}{Qwen3-VL} 
 & DeepSeek-R1     & \textbf{0.760} & \textbf{0.507} & \textbf{0.431} & 0.531 \\
 & GPT-5           & 0.705 & 0.445 & 0.296 & 0.645 \\
 & Qwen3 & 0.734 & 0.470 & 0.361 & 0.494 \\
\midrule
\multirow{3}{*}{GPT-5}
 & DeepSeek-R1     & 0.733 & 0.476 & 0.402 & 0.413 \\
 & GPT-5           & 0.713 & 0.458 & 0.310 & \textbf{0.657} \\
 & Qwen3 & 0.729 & 0.454 & 0.366 & 0.464 \\
\bottomrule
\end{tabular}
\end{table}

We further investigate a multi-agent setup, where the \textit{Knowledge Retrieval Agent} (Retriever) first queries relevant entries from the Knowledge Graph, and the separate \textit{Semantic Reasoning Agent} (Reasoner), instantiated with large language models such as GPT-5, DeepSeek-R1-250528 \citep{Guo2025DeepSeekR1}, or Qwen3-235B-A22B \citep{yang2025qwen3}, subsequently composes the interpretation (Figure~\ref{fig:stage_d}). This separation is motivated by our earlier findings that factual grounding and reasoning fluency benefit from distinct model capabilities.
As shown in Table~\ref{tab:multi_agent}, the multi-agent configurations generally outperform single-agent baselines across the evaluated metrics. We hypothesize that the Semantic Reasoning Agent is better equipped to process and integrate the textual information retrieved from the KG, leveraging its specialized capabilities for enhanced coherence and accuracy. This improvement comes with a moderate increase in inference cost (approximately 1.67× token usage in our profiling).

\subsection{Human experts assessment study} 
To complement the above quantitative metrics, we conducted a human expert evaluation with two Ph.D.~students in archaeology, using the 5-point Likert scale provided in \ref{question}. For fairness, 10\% of the held-out test set was selected, and participants were asked to evaluate the quality of generated interpretations along three pipelines:
(1) the \textit{Baseline pipeline} (direct generation using Qwen3-VL-235B-A22B),
(2) the \textit{RAG pipeline} (retrieval-augmented generation with Qwen3-VL-235B-A22B, and (3) the \textit{Multi-Agent pipeline} (Qwen3-VL-235B-A22B as the Retrieval Agent and DeepSeek-R1-250528 as the Reasoning Agent).

Inter-rater reliability across all annotations was assessed using ICC3 (0.71) and Krippendorff’s Alpha (0.74), indicating substantial agreement between two PhD evaluators with expertise in archaeology. Average Likert scores on a five-point scale showed a clear performance hierarchy. The Multi-Agent Pipeline achieved the highest score of 3.433, followed by the KG-RAG pipeline at 2.133 and the baseline pipeline at 1.367. These human evaluation results are consistent with the automatic metrics and support the reliability of our experimental findings.
To demonstrate the effectiveness of our multi-agent collaborative approach for oracle interpretation, we also qualitatively compare our approach with baseline methods in Figure \ref{fig:instance}.

\section{Supplementary Experiments}

In addition to the main experiments, we further conducted two supplementary studies to test the robustness and generalizability of our approach.

\begin{table}[t]
\centering
\scriptsize
\renewcommand{\arraystretch}{1.1} 
\caption{Results of interpretations conducted in English.}
\label{tab:model_performance_eng}
\setlength{\tabcolsep}{1.5pt} 

\begin{tabular}{@{}lccccc@{}}
\toprule
\rowcolor{header_gray}
\textbf{Category} & \textbf{Model} & \textbf{BERT$\uparrow$} & \textbf{Mover$\uparrow$} & \textbf{ROUGE-1$\uparrow$} & \textbf{LLM-Judge$\uparrow$} \\
\midrule
\multirow{4}{*}{\textit{\textbf{Baseline}}}
& GPT     & 0.152 & -0.123 & 0.111 & 0.098 \\
& Claude  & 0.136 & -0.136 & 0.107 & 0.070 \\
& GLM     & 0.036 & -0.173 & 0.075 & 0.051 \\
& Qwen    & 0.159 & -0.126 & 0.117 & 0.068 \\
\midrule
\multirow{4}{*}{\textit{\textbf{Ours}}}
& GPT     
& 0.272\imp{+0.120}
& 0.034\imp{+0.157}
& 0.201\imp{+0.090}
& \textbf{0.797}\imp{+0.699} \\

& Claude  
& 0.318\imp{+0.182}
& 0.071\imp{+0.207}
& 0.233\imp{+0.126}
& 0.752\imp{+0.682} \\

& GLM     
& 0.318\imp{+0.282}
& \textbf{0.106}\imp{+0.279}
& \textbf{0.268}\imp{+0.193}
& 0.489\imp{+0.438} \\

& Qwen    
& \textbf{0.320}\imp{+0.161}
& 0.075\imp{+0.201}
& 0.234\imp{+0.117}
& 0.619\imp{+0.551} \\
\bottomrule
\end{tabular}
\end{table}

\textbf{English Interpretation Generation.}
To investigate whether the models can generalize across languages, we constructed an English-version task, where the VLMs were required to output interpretations in English rather than Chinese. Results are reported in Table~\ref{tab:model_performance_eng}. Compared with the main Chinese results (Table~\ref{tab:model_performance}), performance is notably lower across all metrics. This degradation is expected, since existing training corpora and retrieval databases are primarily constructed in Chinese, leading to weaker grounding in English. Nevertheless, the relative improvements of retrieval-augmented settings over baseline VLMs remain consistent, suggesting that our pipeline maintains cross-lingual robustness, albeit with a reduced ceiling. These results indicate the importance of developing parallel bilingual resources in paleographic studies to further support cross-linguistic generalization.

\textbf{Variant Character Recognition.}
We evaluate a challenging variant character recognition setting, which requires models to associate visually distinct oracle character variants with a shared canonical form. Performance remains limited across all evaluated models, reflecting the intrinsic difficulty of this task in oracle bone script, where many variants lack explicit component or radical correspondences. Detailed results and expert-informed analysis are provided in Appendix~\ref{app:vari}.

\section{Conclusion}
We propose a component-grounded framework for oracle bone script (OBS) interpretation that leverages the pictographic structure of the script and the relationships among its components. By integrating a component-structured Graph RAG with vision–language models, our approach supports interpretable OBS analysis. We further introduce a component-level oracle dataset and define three progressive tasks, including component retrieval, component relationship inference, and script interpretation, to enable structured evaluation. Experimental results demonstrate that knowledge graph augmentation improves both the accuracy and interpretability of OBS interpretation.
\section*{Limitations}
In collaboration with paleographic experts, we identify several limitations of the current pipeline. Component recognition is not always precise or complete, and the system may occasionally introduce spurious elements. Moreover, a substantial portion of oracle characters still lack widely accepted interpretations, which inherently constrains the reliability of any automated analysis.

Future work may address these limitations by improving component recognition accuracy, expanding the coverage and quality of the underlying knowledge base, and extending the framework to better handle phono-semantic compounds, which remain challenging for current systems.

Finally, our current framework adopts a structured, retrieval-centric workflow rather than fully autonomous generation. This design limits flexibility and relies on external knowledge sources, reflecting the fact that existing VLMs lack intrinsic knowledge of Oracle Bone Script and may hallucinate under unconstrained generation. As base models evolve and acquire stronger domain understanding, future systems may reduce this dependency on explicit retrieval while maintaining philological reliability.

\section*{Ethical Considerations}
This work uses publicly available Oracle Bone Script (OBS) resources and contains no personal, private, or sensitive data. All character- and component-level annotations were conducted by archaeology Ph.D. students with domain expertise in paleography, following authoritative references to ensure accuracy. 

For human evaluation, two Ph.D. students participated voluntarily with informed consent. To reduce fatigue and ensure consistent evaluation conditions, only 10\% of the held-out test set was assessed using a standardized Likert scale.

We acknowledge that automatic interpretation of cultural heritage materials may introduce errors or oversimplifications. Accordingly, the dataset, models, and experimental results presented in this work are intended solely as research aids to support scholarly analysis, and are not designed to replace expert judgment or authoritative paleographic interpretation.

\section*{Acknowledgements}
This work was supported by the National Natural Science Foundation of China (Grant No. 62576148), and the ``Paleography and Chinese Civilization Inheritance and Development Program'' Collaborative Innovation Platform (Grant No. G1917).

\bibliography{custom}

\appendix
\clearpage
\section{Appendix}
\label{sec:appendix}
\subsection{More details on dataset construction}
\label{details_dataset}
\begin{figure}[t]
  \centering
  \includegraphics[width=0.46\columnwidth]{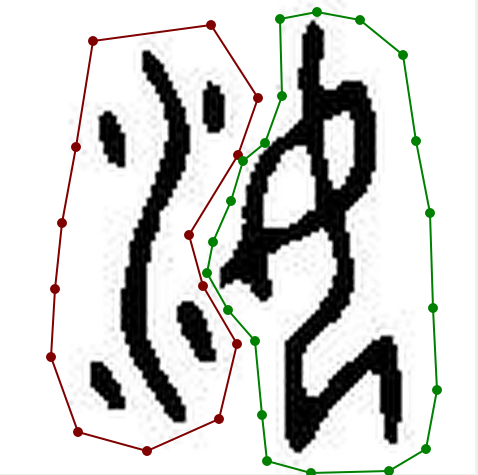}
  \hspace{0.01\columnwidth}
  \includegraphics[width=0.46\columnwidth]{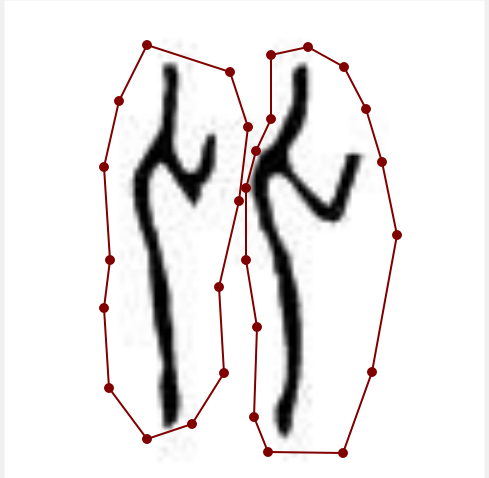}
  \caption{Two images of oracle bone characters segmented by LabelMe.}
  \label{fig:lbimage}
\end{figure}

To ensure fine-grained component-level annotation, we adopted \textbf{LabelMe}\footnote{\url{https://github.com/wkentaro/labelme}} as the primary tool for manual segmentation of Oracle Bone Script images. LabelMe allows annotators to draw polygonal masks directly on images, making it well suited for the irregular shapes and complex outlines of Oracle characters, as shown in Figure \ref{fig:lbimage}.

Each annotation task was conducted by archaeology PhD students who followed authoritative decipherment references. Annotators were compensated for their annotation efforts, with a total payment of 2,450 RMB across all tasks. The process began with drawing precise polygons around each component within a character image. These polygons were then exported into JSON format, which stores the coordinates of the segmentation boundaries together with the corresponding component labels. To improve annotation consistency, we designed a standardized guideline specifying:
\begin{itemize}
    \item \textbf{Segmentation granularity:} ensuring that even small components with distinct semantic functions were delineated separately.
    \item \textbf{Boundary precision:} refining polygon edges to closely follow character contours, especially in cases where strokes overlapped or eroded.
    \item \textbf{Label consistency:} using controlled vocabularies for component names to avoid ambiguity across annotators.
\end{itemize}

As illustrated in Figure~\ref{fig:lbimage}, the annotation workflow produces both the original oracle character and its corresponding component-level masks, which are paired with expert-verified semantic explanations. To ensure annotation reliability, all annotations were performed by archaeology PhD students following authoritative decipherment references, and were subsequently cross-checked to resolve ambiguous boundaries and label inconsistencies.

This expert-curated procedure ensures that \textbf{OB-Radix} achieves high annotation quality and interpretive reliability, laying the foundation for downstream tasks in component recognition and semantic inference.

\newpage
\subsection{Illustration of Component Feature Space Construction}
\label{app:feature_space}

\begin{figure}[t]
  \centering
  \includegraphics[width=0.75\linewidth]{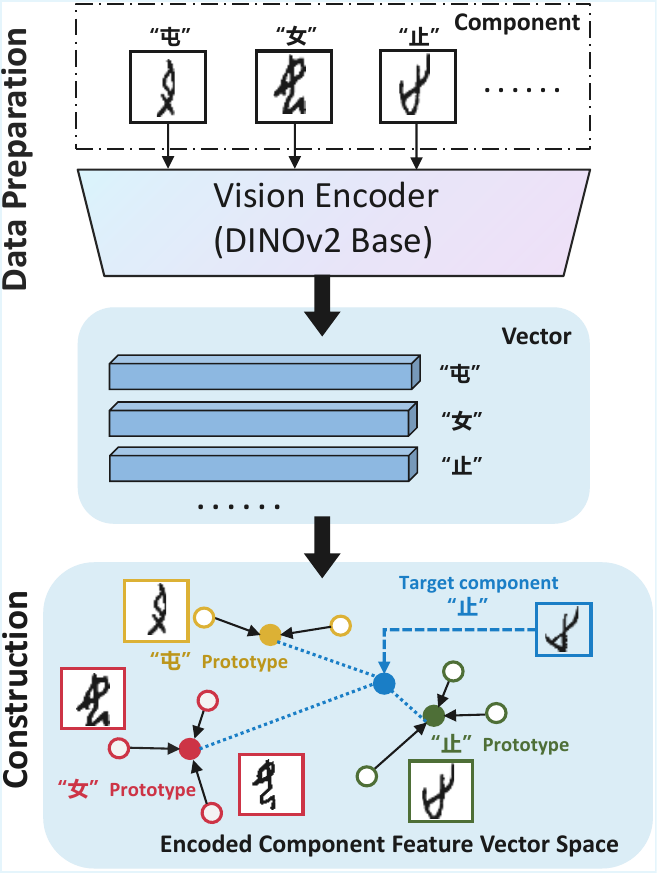}
  \caption{Construction of Vector Space.}
  \label{fig:p4}
\end{figure}

Figure~\ref{fig:p4} provides an intuitive illustration of the component feature space construction process described in Section~\ref{sec:component_identification}.
Each radical image is first encoded by the DINOv2 encoder into a high-dimensional embedding vector.
Images belonging to the same component class form compact clusters in the embedding space, while prototypes (class means) serve as representative anchors for classification.

As shown in the figure, query samples are classified based on their distances to class prototypes rather than individual training instances.
This geometry encourages intra-class compactness and inter-class separability, which is particularly beneficial in low-resource scenarios where only a few labeled examples are available per component.

Such a structured feature space allows the model to generalize effectively to unseen samples while mitigating overfitting, making prototype-based classification well suited for OBS component identification.
\subsection{Oracle Bone Script Interpretation Example}
\label{exp3example}
\begin{figure*}[t]
  \centering
  \includegraphics[width=1\textwidth]{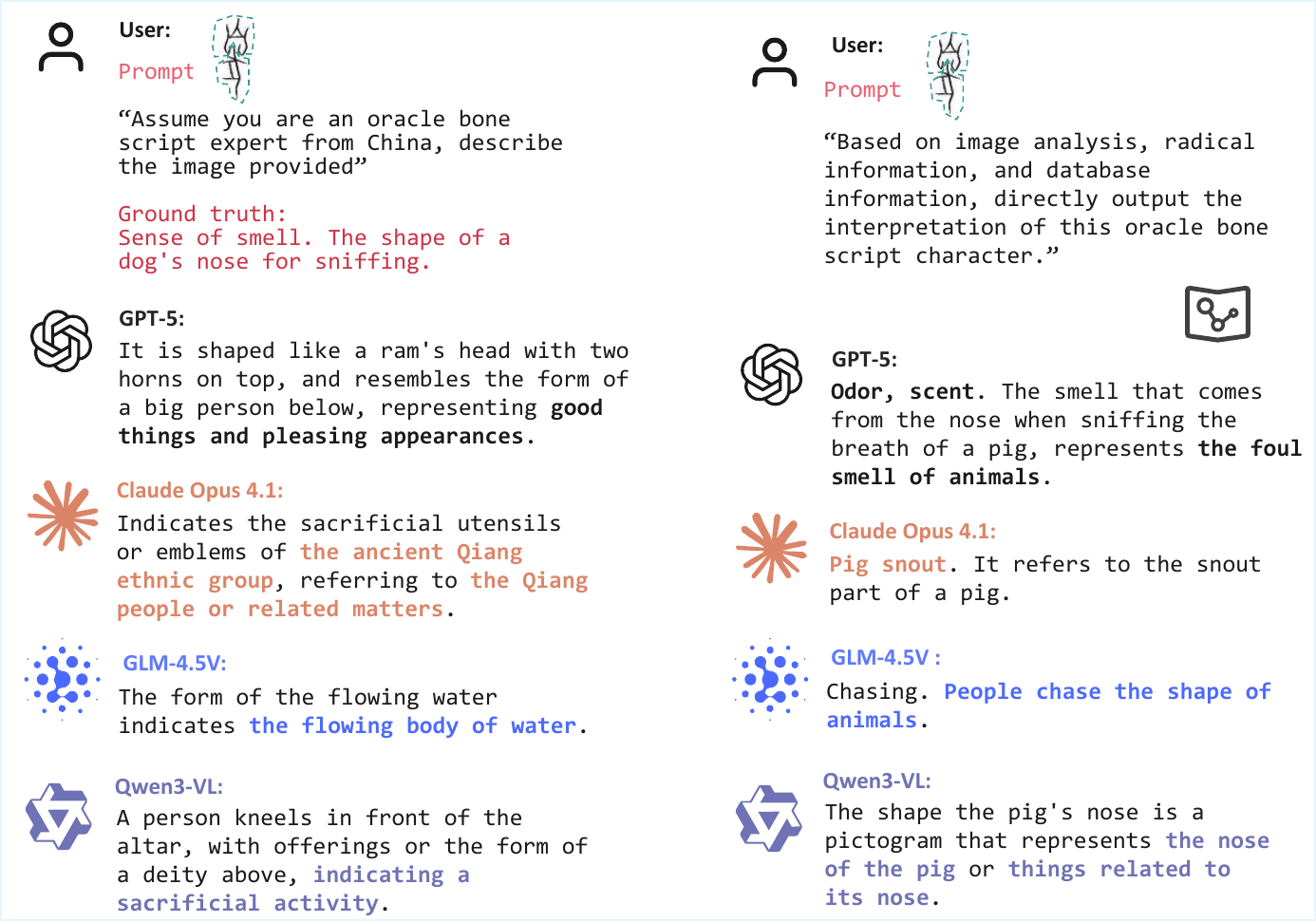}
  \caption{The left side shows the baseline outputs, while the right side shows our results.}
  \label{fig:exp3_example}
\end{figure*}

This section provides an example of oracle bone script (OBS) interpretation generated by our models to illustrate the difference between the \textit{Baseline} and \textit{Agentic RAG} approaches, as shown in Figure~\ref{fig:exp3_example}.

\subsection{LLM-as-a-Judge Evaluation}
\label{app:judge}
To evaluate the semantic correctness of OBS interpretation generation, we adopt an LLM-as-a-Judge evaluation paradigm \citep{zheng2023judging}.
In this setting, a large language model is prompted to compare a model-generated interpretation with a reference interpretation provided by domain experts and to assign a scalar score in the range of 0 to 1, where higher scores indicate better semantic alignment.

The evaluation focuses on semantic consistency rather than surface-level lexical overlap, taking into account the correctness of key entities and participants, core events and relations, semantic modifiers, as well as potential hallucinations or critical omissions.
We instantiate the judge with Gemini~3~Flash using a fixed prompt template and temperature set to zero to ensure deterministic behavior.

\paragraph{Prompt Template.} The exact prompt used for the LLM-as-a-Judge evaluation is shown below:
\begin{tcolorbox}[
    breakable,
    colback=promptbg,      
    colframe=promptborder, 
    boxrule=0.5pt,         
    arc=2mm,               
    left=6pt, right=6pt, top=6pt, bottom=6pt,
    fontupper=\small\sffamily
]
\textbf{System Instruction:}

You are a rigorous semantic assessment expert. You are responsible for scoring the semantic consistency of the sentence to be scored based on the reference sentence.

Scoring Criteria (0.00--1.00): Output a score rounded to the nearest 0.01 (e.g., 0.66, 0.92).

\begin{itemize}\setlength\itemsep{2pt}
  \item \texttt{0.80--1.00 (Perfect)}: Semantically equivalent. Core information and details are accurate, with only reasonable paraphrasing.
  \item \texttt{0.60--0.79 (Excellent)}: Core semantics are accurate. Minor modifiers may be missing, but there are no factual errors.
  \item \texttt{0.40--0.59 (Acceptable)}: Contains key information, but omits some important details or contains minor ambiguities.
  \item \texttt{0.20--0.39 (Poor)}: Significant omission of key information or inclusion of obvious hallucinations, leading to semantic distortion.
  \item \texttt{0.00--0.19 (Failure)}: Completely irrelevant, opposite meaning, or nonsense.
\end{itemize}

\textbf{User Instruction:} Please answer in this format:\\
\texttt{Score: [a number between 0.00 and 1.00]}
\end{tcolorbox}

\subsection{Oracle Bone Script Interpretation Questionnaire}
\label{question}
\begin{figure}[ht]
  \centering
  \includegraphics[width=1\textwidth]{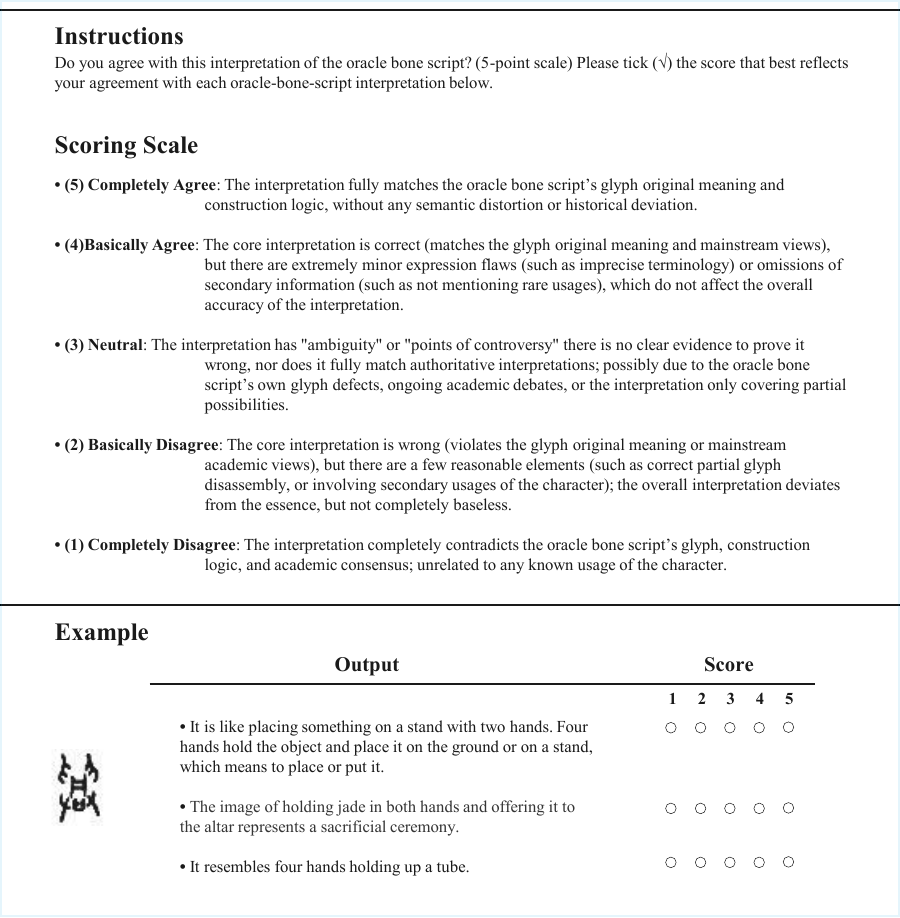}
  \caption{Questionnaire}
  \label{Ques}
\end{figure}

The questionnaire consists of 30 candidate interpretations of oracle bone script characters. Specifically, we curated 10 distinct characters, each of which is associated with three alternative interpretations reflecting different reasoning pipelines. To avoid introducing bias from fixed presentation sequences, the three interpretations corresponding to the same character were randomly permuted prior to distribution. This randomization was applied independently across pipelines, ensuring that participants evaluated the interpretations without being influenced by a consistent order effect. Consequently, the design of the questionnaire facilitates a more balanced and reliable assessment of the comparative quality of the proposed interpretation methods (see Figure~\ref{Ques}).

\subsection{Variant Character Recognition}
\label{app:vari}
Oracle Bone Script contains a large number of variant character forms due to its extended historical usage and diverse engraving practices. Paleographic studies indicate that the script was used for nearly two centuries by multiple engraving groups, resulting in substantial visual variation across characters, often without shared radicals or explicit component structures. To study this phenomenon, we curated 39 variant character pairs and evaluated whether models could associate each variant with its canonical form. As shown in Table~\ref{tab:variant_char}, recognition accuracy remains low across all evaluated models, with no method achieving strong Top-1 or Top-10 performance, reflecting the intrinsic difficulty of this task.

This difficulty arises because many variants lack consistent substructures that can be captured by component- or radical-based visual representations, making compositional cues insufficient in this setting. These results suggest that effective variant recognition may require targeted supervision, explicit variant–canonical mappings, or deeper integration of expert paleographic knowledge, which we leave for future work.

Concretely, one feasible direction is to construct a small but curated set of variant–canonical pairs and perform supervised fine-tuning (SFT) of a VLM to explicitly learn invariances across historically attested shape variations. Such SFT could be combined with deformation-aware augmentation (e.g., stroke-level perturbation, skeleton transformation) to improve robustness to engraving-induced distortions. Another practical approach is to train a contrastive visual encoder where variant–canonical pairs are pulled closer in embedding space, while unrelated characters are pushed apart.
\begin{table}
\centering
\scriptsize
\caption{Variant character search (39 samples).}
\label{tab:variant_char}
\vspace{-6pt}
\setlength{\tabcolsep}{3pt}
\begin{tabular}{@{}lccc@{}}
\toprule
\rowcolor{header_gray}
\textbf{Model} & \textbf{Top-1@ACC} & \textbf{Top-5@ACC} & \textbf{Top-10@ACC} \\
\midrule
GPT-5                 & 5.13\% --- 2 & 5.13\% --- 2 & 5.13\% --- 2 \\
Claude Opus 4.1       & 0.00\% --- 0 & 2.56\% --- 1 & 5.13\% --- 2 \\
GLM-4.5V              & 2.56\% --- 1 & 2.56\% --- 1 & 2.56\% --- 1 \\
Qwen3-VL-235B-A22B    & 2.56\% --- 1 & 2.56\% --- 1 & 5.13\% --- 2 \\
\bottomrule
\end{tabular}
\vspace{-6pt}
\end{table}

\end{document}